%% file: paper.tex
\documentclass[letterpaper]{article} % DO NOT CHANGE THIS
\usepackage{aaai25}  % DO NOT CHANGE THIS
\usepackage{times}  % DO NOT CHANGE THIS
\usepackage{helvet}  % DO NOT CHANGE THIS
\usepackage{courier}  % DO NOT CHANGE THIS
\usepackage[hyphens]{url}  % DO NOT CHANGE THIS
\usepackage{graphicx} % DO NOT CHANGE THIS
\usepackage{natbib}  % DO NOT CHANGE THIS AND DO NOT ADD ANY OPTIONS TO IT
\usepackage{caption} % DO NOT CHANGE THIS AND DO NOT ADD ANY OPTIONS TO IT
\usepackage{algorithm}
\usepackage{algorithmic}

\usepackage{newfloat}
\usepackage{listings}
\DeclareCaptionStyle{ruled}{labelfont=normalfont,labelsep=colon,strut=off} % DO NOT CHANGE THIS
\floatstyle{ruled}
\newfloat{listing}{tb}{lst}{}
\floatname{listing}{Listing}

\usepackage{latexsym}
\usepackage{amssymb}
\usepackage{amsmath}
\usepackage{amsthm}
\usepackage{booktabs}

\usepackage{xspace}
\usepackage{tikz}
\usepackage{xcolor}
\usepackage{enumerate}
\usepackage{bm}
\usepackage{multirow}

\newcommand{\Omit}[1]{}

\newcommand{\set}[1]{\ensuremath{\{ #1 \}}\xspace}
\newcommand{\tup}[1]{\ensuremath{\langle #1 \rangle}\xspace}
\newcommand{\tuple}[1]{\ensuremath{\langle #1 \rangle}\xspace}
\newcommand{\bracket}[1]{\ensuremath{[\![#1]\!]}\xspace}
\newcommand{\multiset}[1]{\ensuremath{\bigl\{\!\!\bigl\{#1\bigr\}\!\!\bigr\}}\xspace}
\renewcommand{\O}{\ensuremath{\mathcal{O}}\xspace}
\newcommand{\citeay}[1]{\citeauthor{#1}~(\citeyear{#1})}
\newcommand{\Q}{\mathcal{Q}}

\newcommand{\baseline}{\text{R-GNN}\xspace}
\newcommand{\extended}[1]{\text{R-GNN}[#1]\xspace}
\newcommand{\allcompositions}{R-GNN$_2$\xspace}
\newcommand{\edgetransformer}{ET\xspace}
\newcommand{\twognn}{2-GNN\xspace}
\newcommand{\A}{\ensuremath{\mathcal{A}}\xspace}

\newcommand{\norm}{\text{LN}}
\newcommand{\triatt}{\text{Tri.Att.}}
\newcommand{\mlp}{\text{MLP}}
\newcommand{\comb}{\text{comb}}
\newcommand{\agg}{\text{agg}}
\newcommand{\join}{\triangle}

\newcommand{\STRUC}{\ensuremath{\mathsf{STRUC}}\xspace}

\newcommand{\J}{\ensuremath{\mathcal{J}}\xspace}
\newcommand{\D}{\ensuremath{\mathcal{D}}\xspace}

\newtheorem{definition}{Definition}
\newtheorem{theorem}[definition]{Theorem}

\let\oldequation\equation
\let\endoldequation\endequation
\renewenvironment{equation}{\footnotesize\oldequation}{\endoldequation}

\let\oldalign\align
\let\endoldalign\endalign
\renewenvironment{align}{\footnotesize\oldalign}{\endoldalign}

\title{Learning More Expressive General Policies for Classical Planning Domains}
\author{
    Simon St\r{a}hlberg\textsuperscript{\rm 1},
    Blai Bonet\textsuperscript{\rm 2},
    Hector Geffner\textsuperscript{\rm 1}
}
\affiliations{
    \textsuperscript{\rm 1}RWTH Aachen University, Germany\\
    \textsuperscript{\rm 2}Universitat Pompeu Fabra, Spain\\
    simon.stahlberg@gmail.com, bonetblai@gmail.com, hector.geffner@ml.rwth-aachen.de
}

\nocopyright{}

\begin{document}

\maketitle

\begin{abstract}
  \input{sections/abstract.tex}
\end{abstract}

\input{sections/introduction.tex}
\input{sections/related_work.tex}
\input{sections/background.tex}
\input{sections/gnns.tex}
\input{sections/relational_gnns.tex}
\input{sections/weisfeiler_leman.tex}
\input{sections/extended_relational_gnns.tex}
\input{sections/learning.tex}
\input{sections/baselines.tex}
\input{sections/example.tex}
\input{sections/experiments.tex}
\input{sections/expressivity.tex}
\input{sections/conclusions.tex}
\input{sections/acknowledgements.tex}

\bibliography{../bibliography,../crossref-abbrv}

\end{document}

%% file: sections/abstract.tex
GNN-based approaches for learning general policies across planning domains are
limited by the expressive power of $C_2$, namely; first-order logic with two
variables and counting. This limitation can be overcame by transitioning to
$k$-GNNs, for $k\,{=}\,3$, wherein object embeddings are substituted with
triplet embeddings. Yet, while $3$-GNNs have the expressive power of $C_3$,
unlike $1$- and $2$-GNNs that are confined to $C_2$, they require quartic time
for message exchange and cubic space to store embeddings, rendering them
infeasible in practice.
In this work, we introduce a parameterized version \extended{$t$}
(with parameter $t$) of Relational GNNs. Unlike GNNs, that are designed to
perform computation on graphs, Relational GNNs are designed to do computation on
relational structures. When $t\,{=}\,\infty$, \extended{$t$} approximates
$3$-GNNs over graphs, but  using only quadratic space for embeddings. For lower
values of $t$, such as $t\,{=}\,1$ and $t\,{=}\,2$, \extended{$t$} achieves a
weaker approximation by exchanging fewer messages, yet interestingly, often
yield the expressivity required in several planning domains. Furthermore, the
new \extended{$t$} architecture is the original \baseline architecture with a
suitable transformation applied to the inputs only. Experimental results
illustrate the clear performance gains of \extended{$1$} over
the plain {\baseline}s, and also over Edge Transformers that also approximate
$3$-GNNs.

%% file: sections/introduction.tex
\section{Introduction}

General policies are policies that can be used to solve a collection of planning
problems
reactively~\cite{srivastava-et-al-aaai2008,hu-degiacomo-ijcai2011,belle-levesque-kr2016,bonet-geffner-ijcai2018,illanes-mcilraith-aaai2019,jimenez-et-al-ker2019}.
For example, a general policy for solving all Blocksworld problems can place all
blocks on the table, and then build up the target towers from the bottom up. Yet
while nearly perfect general policies have been learned for many classes of
planning
domains~\cite{toyer-et-al-jair2020,rivlin-et-al-icaps2020wsprl,stahlberg-et-al-icaps2022},
one key expressive limitation results from the type of features used to
classify state transitions or actions. In combinatorial approaches, features are
selected from a domain-independent pool, created using a description logic
grammar~\cite{baader-et-al-2003} based on the given domain
predicates~\cite{bonet-geffner-ijcai2018,bonet-et-al-aaai2019}, while in deep
learning approaches, the features are learned using relational versions of graph
neural networks
\cite{scarselli-et-al-ieeenn2009,gilmer-et-al-icml2017,hamilton-2020}. A shared
limitation of \emph{both} approaches, however, is their inability to learn
policies requiring complex logical features. This limitation arises in
description logics from the $C_2$ fragment of first-order logic that they
capture; namely, first-order logic limited to two variables and
counting~\cite{baader-et-al-2003}, and in GNNs, from the type of message passing
that is accommodated, where direct communication involves pairs of objects but
no triplets~\cite{grohe-lics2021}.

This expressive limitation, not always acknowledged, is serious. For example,
although these methods can learn general policies for guiding an agent to a
specific cell in an $n \times n$ grid containing \emph{obstacles}, with
positions and adjacency relations defined in terms of cells and atoms such as
$\textsc{At}(c)$ and $\textsc{Adj}(c,c')$, they lack the \emph{expressive
capacity} when the relations are represented with atoms like $\textsc{At}(x,y)$,
$\textsc{Adj}_1(x, x')$, and $\textsc{Adj}_2(y, y')$. Similarly, these methods
are unable to learn policies for classical benchmark domains such as Logistics
and Grid, that require composition of binary relations, which is beyond the
scope of $C_2$ \cite{stahlberg-et-al-kr2022,stahlberg-et-al-kr2023}.

In principle, this limitation can be addressed by using richer grammars to
generate non-$C_2$ features, in the logical setting, or by using $k$-GNNs, with
$k\,{=}\,3$, on the neural setting, where triplets of objects are embedded
instead of individual objects~\cite{morris-et-al-aaai2019}. It is known that
3-GNNs have the expressive power of $C_3$ logic, unlike the $C_2$ expressive
power of 1- and 2-GNNs~\cite{grohe-lics2021}. Yet 3-GNNs do not scale up as they
require cubic number of embeddings, and quartic time for exchanging messages.

In this paper, we introduce an alternative, parameterized version of Relational
GNNs (R-GNNs). R-GNNs are designed to perform computation over relational
structures, unlike GNNs that can only process graphs. The architecture for
\extended{$t$} mirrors that of plain R-GNNs and differs only in the input. While
a plain R-GNNs takes the set of atoms $S$ representing a planning state as
input, \extended{$t$} accepts a transformed set of atoms $A_t(S)$ instead. At
$t\,{=}\,0$, \extended{$t$} approximates 3-GNNs weakly, while at
$t\,{=}\,\infty$, it offers a strong approximation. Thus, the parameter $t$
serves to balance expressive power with computational effort. Crucially, for
lower values of $t$, such as $t\,{=}\,1$ and $t\,{=}\,2$, \extended{$t$}'s
message passing runs in quadratic time in general while capturing the $C_3$
features that are essential in several planning domains. Our experiments
demonstrate that \extended{$t$}, even with small values of $t$, is practically
feasible and significantly improves both the coverage and the quality of the
learned general plans when compared to four baselines: plain \baseline{},
2-GNN, \allcompositions{}, and Edge-Transformers~\cite{bergen-et-al-neurips2021},
where the last two aim to approximate 3-GNNs~\cite{bergen-et-al-neurips2021} more
accurately than \extended{$t$}.

The rest of the paper is organized as follows. We review first related work and
background on planning, generalized planning, GNNs and relational GNNs,
and the Weisfeiler-Leman coloring algorithms.
Then we introduce the parametric R-GNNs, the learning task and baselines, and the
experimental results. The paper ends with a discussion on the expressivity of 
the model, and conclusions.

%% file: sections/related_work.tex
\section{Related Work}

\paragraph{General policies from logic.}
The problem of learning general policies has a long
history~\cite{khardon-aij1999,martin-geffner-ai2004,fern-et-al-jair2006}, and
general policies have been formulated in terms of
logic~\cite{srivastava-et-al-aij2011,illanes-mcilraith-aaai2019},
regression~\cite{boutilier-et-al-ijcai2001,wang-et-al-jair2008,sanner-boutelier-aij2009},
and policy rules~\cite{bonet-geffner-ijcai2018,bonet-et-al-aaai2019} that can
be learned \cite{frances-et-al-aaai2021,drexler-et-al-icaps2022}.

\paragraph{General policies from neural nets.} Deep learning (DL) and deep
reinforcement learning (DRL)
\cite{sutton-barto-1998,bertsekas-1995,vincent-et-al-ftml2018} have been used to
learn general policies~\cite{kirk-et-al-jair2023}. In some cases, the planning
representation of the domains is
used~\cite{toyer-et-al-jair2020,bajpai-et-al-neurips2018,rivlin-et-al-icaps2020wsprl};
in most cases, it is
not~\cite{groshev-et-al-icaps2018,chevalier-boisvert-et-al-iclr2019}, and in
practically all cases, the neural networks are GNNs or variants. Closest to our
work is the use of GNNs for learning general policies for classical
planning~(\citeauthor{stahlberg-et-al-kr2022},~\citeyear{stahlberg-et-al-kr2022,stahlberg-et-al-kr2023}).

\paragraph{GNNs, R-GNNs, and $C_k$ logics.} The use of GNNs is common when
learning general policies where the number of objects change from instance to
instance. This is because GNNs trained with small graphs can be used for dealing
with larger
graphs~\cite{scarselli-et-al-ieeenn2009,gilmer-et-al-icml2017,hamilton-2020},
and because states in classical planning are closely related to graphs: they
represent relational structures that become graphs when there is a single
non-unary relation that is binary and symmetric. In such a case, the graph
vertices stand for the objects and the edges for the relation. Relational GNNs
extend GNNs to relational
structures~\cite{schlichtkrull-et-al-eswc2018,vashishth-et-al-iclr2019,barcelo-et-al-lgc2022},
and our R-GNNs borrow from those used for max-CSP~\cite{toenhoff-et-al-ewsp2021}
and generalized planning~\cite{stahlberg-et-al-icaps2022}.

There is a tight correspondence between the classes of graphs that can be
distinguished by GNN, the WL
procedure~\cite{morris-et-al-aaai2019,xu-et-al-iclr2019}, and $C_2$
logic~\cite{cai-furer-immerman-combinatorica1992,barcelo-et-al-iclr2020,grohe-lics2021}.
The expressive power of GNNs can be extended by replacing graph vertices by
tuples of $k$-vertices. The resulting $k$-GNNs have the the power of the $k$-WL
coloring algorithm, and hence the expressivity of $C_{k}$ for $k > 2$. The
``folklore'' variant of the $k$-WL algorithm,
$k$-FWL~\cite{cai-furer-immerman-combinatorica1992}, is more efficient as it has
the power of $(k\!+\!1)$-WL while using $\O(n^k)$ memory.
\citeay{maron-et-al-iclr2019} define a parameterized family of
permutation-invariant neural networks, which for $k=2$ has the expressiveness of
$3$-GNNs~\cite{maron-et-al-neurips2019}. In the experiments, we consider a
baseline based on Edge Transformers that has the same
expressiveness~\cite{bergen-et-al-neurips2021}.

%% file: sections/background.tex
\section{Background}

We review planning, generalized planning, GNNs, relational GNNs, and the Weisfeiler-Leman graph coloring algorithms.

\subsection{Planning and Generalized Planning}

A classical planning problem is a pair $P\,{=}\,\tup{D, I}$, where $D$
represents a first-order \emph{domain} and $I$ contains information specific to
the \emph{problem
instance}~\cite{ghallab-et-al-2004,geffner-bonet-2013,haslum-et-al-2019}. The
domain $D$ mainly consists of two components: a set of predicate symbols, and a set of
action schemas. The action schemas come with preconditions and effects expressed
with atoms $p(x_1, x_2, \ldots, x_k)$ where $p$ is a predicate symbol (also called domain predicate) of arity $k$,
and each term $x_i$ is a schema argument. An instance is a tuple
$I\,{=}\,\tup{O, S_I, G}$, where $O$ represents a set of object names, $S_I$ is the
initial state expressed as a set of \emph{ground atoms} $p(o_1, o_2, \ldots, o_k)$,
where $o_i \in O$ and $p$ is a predicate of arity $k$, and $G$ is also a set of ground
atoms encoding the goal. A problem $P$ compactly defines
a transition system over a finite set of states.

A \emph{generalized policy} $\pi$ for a class $\Q$ of planning instances over
the same domain $D$ represents a collection of state transitions $(S,S')$ in
each instance $P$ of $\Q$ that are said to be in $\pi$. A $\pi$-trajectory is a
sequence of states $S_0, S_1, \ldots, S_n$ that starts in the initial state of $P$
and whose transitions $(S_i,S_{i+1})$ are all in $\pi$. The trajectory is
maximal if $S_n$ is the first goal state of the sequence or there is no
transition $(S_n,S)$ in $\pi$. The policy $\pi$ solves $P$ if all maximal
$\pi$-trajectories in $P$ reach the goal, and it solves $\Q$ if it solves each
$P$ in $\Q$.
A policy $\pi$ can be represented in many forms from formulas or rules to general
value functions $V(S)$; in the latter, the state transitions $(S,S')$ in $\pi$
are those that minimize the value $V(S')$, for the successor states $S'$ of $S$.

%% file: sections/gnns.tex
\subsection{Graph Neural Networks (GNNs)}

GNNs are parametric functions that operate on graphs
\cite{scarselli-et-al-ieeenn2009,gilmer-et-al-icml2017,hamilton-2020}. GNNs
maintain and update embeddings $\bm{f}_i(v)\,{\in}\,\mathbb{R}^k$ for each vertex
$v$ in a graph $G$. The process is iteratively performed over $L$ layers, from
initial embeddings $\bm{f}_0(v)$, and progressing for $i=0,\ldots,L-1$:
\begin{equation}
  \bm{f}_{i+1}(v) = \comb{}_i\bigl( \bm{f}_i(v), \agg{}_i\bigl(\multiset{ \bm{f}_i(w) \mid w\,{\in}\,N_G(v)} \bigr) \bigr)
  \label{eq:gnn:update}
\end{equation}
where $\agg{}_i$ and $\comb{}_i$ are aggregation and combination functions,
respectively, and
\multiset{ \bm{f}_i(w) \mid w\,{\in}\,N_G(v)} is the \emph{multiset}
of embeddings $\bm{f}_i(w)$ for the neighboring vertices $w$ of $v$ in the
graph $G$.
The aggregation functions $\agg{}_i$ (e.g., max, sum, or smooth-max)
condense multiple vectors into a single vector, whereas the combination
functions $\comb{}_i$ merge pairs of vectors. The function implemented by GNNs
is well defined for graphs of any size, and invariant, under (graph)
isomorphisms, for permutation-invariant aggregation functions. The aggregation
and combination functions are parametric, allowing the vertex embeddings
$\bm{f}_i(\cdot)$ to be learnable functions.

%% file: sections/relational_gnns.tex
\begin{algorithm}[t]
    \caption{Relational GNN (R-GNN)}
    %\resizebox{\linewidth}{!}{
    \begin{minipage}{1.1\linewidth}
    \begin{algorithmic}[1]
        \STATE \textbf{Input:} Set of ground atoms $S$ (state), and objects $O$
        \STATE \textbf{Output:} Embeddings $\bm{f}_L(o)$ for each object $o \in O$
        \STATE Initialize $\bm{f}_0(o) \sim 0^k$ for each object $o \in O$
        \FOR{$i \in \{0, \ldots, L-1\}$}
        \FOR{each atom $q := p(o_1, o_2, \ldots, o_m) \in S$}
        \STATE $\bm{m}_{q,o_j} :=\, [\mlp{}_p(\bm{f}_i(o_1), \bm{f}_i(o_2), \ldots, \bm{f}_i(o_m))]_j$ %, 1 \leq j \leq m$
        \ENDFOR
        \FOR{each object $o \in O$}
        \STATE $\bm{f}_{i+1}(o) \,:=\, \bm{f}_{i}(o)$
        \STATE \hspace{1em} $\,{+}\,\mlp{}_U\bigl(\bm{f}_i(o), \agg{}\bigl(\multiset{\bm{m}_{q,o}\,{\mid}\, o\,{\in}\,q,q\,{\in}\,S}\bigr)\bigr)$
        \ENDFOR
        \ENDFOR
    \end{algorithmic}
    \end{minipage}
    %}
    \label{alg:relnn}
\end{algorithm}

\subsection{Relational GNNs (R-GNNs)}

GNNs operate over graphs, whereas planning states are relational structures over
predicates of varying arities. The Relational GNN (R-GNN) for processing
relational structures \cite{stahlberg-et-al-icaps2022} is inspired by those used
for max-CSPs~\cite{toenhoff-et-al-ewsp2021}. Like GNNs, R-GNNs is a
message-passing architecture where messages are exchanged between the objects in
the input relational structure \A, but the messages are not directly associated
with edges. Instead, messages are exchanged according to the atoms that are true
in the structure. That is, initial embeddings $\bm{f}_0(o)$ for each object $o$
in \A are updated as follows:
\begin{equation}
  \bm{f}_{i+1}(o) = \comb{}_i\bigl(\bm{f}_i(o), \agg{}_i\bigl(\multiset{\bm{m}_{q,o} \mid o \in q, \A\vDash q}\bigr)\bigr) \,,
  \label{eq:rnn:update}
\end{equation}
where $\bm{m}_{q,o}$ is the message that atom $q$ (that is true in the
relational structure \A, and that mentions object $o$) sends to object $o$. For
a $m$-ary predicate $p$, an atom of the form $q\,{=}\,p(o_1,o_2,\ldots,o_m)$
sends $m$ (non-necessarily equal) messages to the objects $o_1,o_2,\ldots,o_m$,
respectively. All such messages are computed (in parallel) using a
\emph{learnable} combination function $\comb{}_p(\cdot)$, one for each symbol
$p$, that maps $m$ input embeddings into $m$ output messages:
\begin{equation}
  \bm{m}_{q,o_j} = \bigl[\comb{}_p\bigl(\bm{f}_i(o_1), \bm{f}_i(o_2), \ldots, \bm{f}_i(o_m)\bigr)\bigr]_j \,,
\end{equation}
where $[\ldots]_j$ refers to the $j$-th embedding of its argument.
The $\comb{}_i(\cdot)$ function in \eqref{eq:rnn:update} merges two vectors of
size $k$, the current embedding $\bm{f}_i(o)$ and the aggregation of the
messages received at object $o$.

The relational neural network for planning states $S$ is detailed in
Algorithm~\ref{alg:relnn}, where the update for the embeddings in
\eqref{eq:rnn:update} is implemented via \emph{residual connections.} In our
implementation, the aggregation function $\agg{}(\cdot)$ is \emph{smooth
maximum} that approximates the (component-wise) maximum. The combination
functions are implemented using MLPs. The functions $\comb{}_i(\cdot)$
correspond to the same $\mlp{}_U$ that maps two real vectors in $\mathbb{R}^k$
into a vector in $\mathbb{R}^k$, while $\comb{}_p(\cdot)$, for a predicate $p$
of arity $m$, is an $\mlp{}_p$ that maps $m$ vectors in $\mathbb{R}^k$ into $m$
vectors in $\mathbb{R}^k$. In all cases, each $\mlp{}$ has three parts: first, a
linear layer; next, the Mish activation function~\cite{misra-bmva2020}; and then
another linear layer. The architecture in Algorithm~\ref{alg:relnn} requires two
inputs: a set of atoms denoted as $S$, and a set of objects denoted as $O$. The
goal $G$ is encoded by \emph{goal atoms} that are assumed to be in $S$: if
$p(o_1, o_2, \ldots, o_m)$ is an atom in $G$, the atom $p_g(o_1, o_2, \ldots,
o_m)$ is added to $S$, where $p_g$ is a new ``goal predicate''
\cite{martin-geffner-ai2004}.

The set of object embeddings $\bm{f}_L(o)$ at the last layer is the result of
the net; i.e., $\text{R-GNN}(S,O)\,{=}\,\multiset{ \bm{f}_L(o) \mid o\in O }$.
Such embeddings are used to encode general value functions, policies, or both.
In this work, we encode a learnable value function $V(S)$ through a simple
additive readout that feeds the embeddings into a final MLP:
\begin{equation}
  V(S) = \mlp{}\bigl(\textstyle\sum_{o \in O} \bm{f}_L(o)\bigr) \,.
\end{equation}

%% file: sections/weisfeiler_leman.tex
\subsection{Weisfeiler-Leman Coloring Algorithms}

Weisfeiler-Leman (WL) coloring algorithms provide the theory
for establishing the expressive limitation of GNNs.
These algorithms iteratively color each vertex of a graph, or each $k$-tuple
of vertices, based on the colors of their neighbors, until a fixed point is reached.
Colors, represented as natural numbers, are generated
with a \textsc{Relabel} function that maps structures over colors into unique
colors. We borrow notation and terminology from \citeay{morris-et-al-jmlr2023}.

\paragraph{1-Dimensional WL (1-WL).}
For a  graph $G\,{=}\,(V, E)$, the node coloring $C^1_{i + 1}$ at iteration $i$ is
defined as:
\begin{equation}
    C^1_{i + 1}(v) = \textsc{Relabel}\bigl(\bigl\langle C^1_{i}(v), \multiset{C^1_{i}(u) \mid u \in N(v)} \bigr\rangle\bigr)
\end{equation}
where $N(v)$ is the neighborhood of node $v$, and the initial coloring is
determined by the given vertex colors, if any, or uniform otherwise.

\paragraph{Folklore $k$-Dimensional WL ($k$-FWL).}
For a graph $G$ and tuple $\tuple{v} \in V(G)^k$, the coloring $C^k_{i + 1}$ at
iteration $i$ is:
\begin{equation}
    C^k_{i+1}(\tuple{v}) = \textsc{Relabel}\bigl(\bigl\langle C^k_i(\tuple{v}), M_i(\tuple{v}) \bigr\rangle\bigr)
\end{equation}
where $M_i(\tuple{v})$ is the multiset of tuples
\begin{equation}
    \begin{split}
        M_i(\tuple{v}) = \multiset{\bigl(&C^k_i(\phi_1(\tuple{v}, w)),\dots,\\
        &\quad C^k_i(\phi_k(\tuple{v}, w))\bigr) \mid w\,{\in}\,V(G)} \,,
    \end{split}
\end{equation}
the function $\phi_j(\tuple{v}, w)$ replaces the $j$-th component of the tuple
$\tuple{v}$ with the node $w$, and the initial color for tuple $\tuple{v}$ is
determined by the structure of the subgraph induced by $\tuple{v}$, and the
order of the vertices in the tuple $\tuple{v}$.

\paragraph{Oblivious $k$-Dimensional WL ($k$-OWL).}
The coloring $C^{k*}_{i + 1}$ for the $k$-OWL variant at iteration $i$ is defined as:
\begin{equation}
    C^{k*}_{i+1}(\tuple{v}) = \textsc{Relabel}\bigl(\bigl\langle C^{k*}_i(\tuple{v}), M^*_i(\tuple{v}) \bigr\rangle\bigr)
\end{equation}
where the multiset $M_i(\tuple{v})$ of vectors in $k$-FWL is replaced by a
$k$-dimensional vector $M^*_i(\tuple{v})$ of multisets:
\begin{equation}
    \bigl[ M^*_i(\tuple{v}) \bigr]_j = \multiset{C^{k*}_i(\phi_j(\tuple{v}, w)) \mid w \in V(G)}
\end{equation}
for $j=1,2,\ldots,k$, where $\phi_j(\tuple{v}, w)$ is the same function as in
$k$-FWL, and the initial coloring $M^*_0(\tuple{v})$ is also the same.

For a tuple $\tuple{v}$ of $k$ vertices, there are potentially $k \cdot n$
neighbor-tuples resulting from replacing each $j$-th component $v_j$ of
$\tuple{v}$ with each of the $n$ nodes in the graph, for $j\,{=}\,1,2,\ldots,
k$. The key difference between $k$-FWL and $k$-OWL lies in how these $k \cdot n$
tuples are grouped to determine the new color of $\tuple{v}$. In $k$-FWL, the
tuple $\tuple{v}$ ``sees'' a multiset of $n$ vectors, each with $k$ colors,
resulting from replacing $v_j$ with $w$ for each $j$ from 1 to $k$, providing
one such vector or ``context'' for each node $w$ in the graph. In contrast, in
$k$-OWL, the tuple $\tuple{v}$ ``sees'' a $k$-vector whose elements
$\tuple{v}_j$ are multisets of $n$ colors, with the $j$ component $v_j$ of
$\tuple{v}$ replaced by each of the $n$ nodes in the graph. In $k$-OWL, the
``contexts'' mentioned above are broken, hence the method is termed
``oblivious'' as it is oblivious to such contexts.

\paragraph{2-FWL vs.\ 2-OWL.}
The case for $k\,{=}\,2$ clearly illustrates the difference between the two
algorithms. For a graph $G\,{=}\,(V,E)$ and coloring $C$, the context for the
pair \tup{u_,v} considered by 2-FWL is
{\footnotesize % TODO.
\begin{equation*}
  M(\tup{u,v})\,{=}\,\multiset{ \bigl(C(\tup{w,v}), C(\tup{u,w})\bigr) \mid w \in V }\,.
\end{equation*}
}%
It considers pairs of tuples \tup{u,w} and \tup{w,v} whose ``join'' results in
\tup{u,v}. On the other hand, the context considered by 2-OWL is
{\footnotesize % TODO.
\begin{equation*}
  M^*(\tup{u,v})\,{=}\,\bigl(\multiset{C(\tup{w,v})\,{\mid}\,w\,{\in}\,V}, \multiset{C(\tup{u,w})\,{\mid}\,w\,{\in}\,V}\bigr).
\end{equation*}
}%
It is clear the ``loss of information'' suffered by 2-OWL with respect to 2-FWL
as the context $M^*(\tup{u,v})$ can be recovered from %the context
$M(\tup{u,v})$, but not the other way around.

\paragraph{Expressive power.}
\citeay{cai-furer-immerman-combinatorica1992} established that two graphs are
indistinguishable by $k$-FWL if and only if they satisfy the same set of
formulas in the logic $C_{k + 1}$. In terms of expressive power, 1-OWL is
equivalent to 2-OWL, $k$-OWL is strictly more expressive than $(k-1)$-OWL, for
$k\geq 3$, and $k$-OWL has the same expressiveness as $(k-1)$-FWL, for $k \geq
2$. More importantly, the discriminative power of $C_3$ can be achieved either
by using $3$-OWL over triplets in cubic space, or by using $2$-FWL over pairs in
quadratic space.

%% file: sections/extended_relational_gnns.tex
\section{Parametric Extended R-GNN: R-GNN[$t$]}

The new architecture extends the expressive power of R-GNNs beyond $C_2$ by
capitalizing the relational component of R-GNNs, outlined in
Algorithm~\ref{alg:relnn}. Indeed, the function computed by the new
architecture, $\text{R-GNN}[t](S,O)$, where $t$ is a non-negative integer
parameter, is defined as:
\begin{equation}
    \hbox{R-GNN}[t](S,O) = \hbox{R-GNN}(A_t(S),O^2)
    \label{rnnt}
\end{equation}
where $O^2\,{=}\,O\,{\times}\,O$ stands for the pairs $\tup{o, o'}$ of objects
in $O$, and $A_t(S)$ stands for a transformation of the atoms in $S$ that
depends on the parameter $t$. Specifically, if $w\,{=}\,\tup{o_1, o_2, \ldots,
o_m}$ is a tuple of objects, $\tup{w}^2$ refers to the tuple of $m^2$ pairs
obtained from $w$ as:
\begin{equation}
    \tup{w}^2 = \tup{(o_1,o_1), \ldots, (o_1,o_m), \ldots, (o_m,o_1),\ldots,(o_m,o_m)}
\end{equation}
Then, for $t\,{=}\,0$, the set of atoms $A_t(S)$ is:
\begin{equation}
    A_0(S) = \set{ p(\tup{w}^2)  \mid  p(w) \in S } \,.
\end{equation}
That is, predicates $p$ of arity $m$ in $S$ transform into predicates $p$ of
arity $m^2$ in $A_0(S)$, and each atom $p(w)$ in $S$ is mapped to the atom
$p(\tup{w}^2)$.

For $t\,{>}\,0$, the set of atoms $A_t(S)$ extends $A_0(S)$ with atoms for a new
ternary predicate $\join$ as: $A_t(S)\,{=}\,A_0(S) \cup \Delta_t(S)$ where
\begin{equation}
    \Delta_t(S) = \bigl\{\join(\tup{o,o'}, \tup{o',o''},\tup{o,o''}) \mid \tup{o, o'}, \tup{o', o''} \in R_t \bigr\} \,,
\end{equation}
and the binary relation $R_t$ is defined from $S$ and $G$ as:
\begin{equation}
    \langle o, o' \rangle \in R_t \text{ iff }
    \begin{cases}
      \text{$o$ and $o'$ are both in an atom in $S$}              & \text{if $t=1$,} \\
      \text{$\exists o''[\set{\tup{o,o''}, \tup{o'',o'}} \subseteq R_{t-1}]$} & \text{if $t>1$.}
    \end{cases}
\end{equation}
In words, for the R-GNN to emulate a relational version of 2-FWL, two things are
needed. First, object pairs need to be embedded. This is achieved by replacing
the atoms in $S$ with the atoms in $A_0(S)$ whose arguments are object pairs.
Second, object pairs $\tup{o,o'}$ need to receive and aggregate messages from
object triplets, the ``contexts'' in 2-FWL, which are formed by vectors of pairs
$\tup{o,o''}$ and $\tup{o'',o'}$. This interaction is captured through the new
atoms $\join(\tup{o,o''},\tup{o'',o'},\tup{o,o'})$ and the associated
$\mlp_{\join}$. The relational GNN architecture in Algorithm~\ref{alg:relnn}
allows each argument to communicate with every other argument in the context of
a third one, with messages that depend on all the arguments. This is similar to
the ``triangulation'' found in the Edge Transformer. But, rather than adding all
possible $\join(\tup{o,o''},\tup{o'',o'},\tup{o,o'})$ atoms to $A_0(S)$, the
atoms are added in a controlled manner using the parameter $t$ to avoid a cubic
number of messages to be exchanged. The parameter $t$ controls the maximum
number of sequential compositions that can be captured. In problems that require
a single composition, a value of $t\,{=}\,1$ suffices to yield the necessary
$C_3$ features without having to specify which relations need to be composed.
Moreover, all this is achieved by simply changing the input from \tup{S,O} to
\tup{A_t(S),O^2}.

The final embeddings produced by \extended{$t$} are then used to define a
general value function $V(S)$ as
\begin{equation}
  V(S) = \mlp{}\bigl(\textstyle\sum_{o \in O} \bm{f}_L(\langle o, o \rangle)\bigr) \,,
  \label{eq:final}
\end{equation}
where the readout only takes the final embeddings for object pairs $\tup{o,o}$
that represent single objects, and passes their sum to an MLP that outputs the
scalar $V(S)$. The reason is to avoid summing the embeddings for all pairs
$\tup{o,o'}$ as it leads to high variance, and a more difficult learning.

Since \extended{$t$} is a regular \baseline over a transformed input, the
objects in a {\baseline} are indistinguishable a priori, and the readout
function only considers pairs of type $\tup{o,o}$, some way to make such pairs
different from others must be incorporated so that the message passing mechanism
learns where to send the information for the readout. This is automatically
achieved by adding static atoms $\textsc{Obj}(o)$, for each object $o$, for a
new static unary predicate $\textsc{Obj}$. Such atoms are then transformed into
atoms $\textsc{Obj}(\tup{o,o})$ in $A_0(S)$ that mark the pairs $\tup{o,o}$ as
different from other pairs $\tup{o,o'}$.

%% file: sections/learning.tex
\section{Learning Task}

We aim at extending the expressivity of R-GNNs for finding policies for
generalized planning. For this purpose, for each domain in the benchmark, we
learn a general value function $V$ in a supervised manner from the optimal
values $V^*(S)$ over a small collection of training states $S$. Such states $S$
belong to planning instances with small state spaces which makes the computation
of $V^*$ straightforward with a standard breadth-first search. The loss function
that is minimized during training is:
\begin{equation}
    L(S) = |V^*(S) - V(S)| \,.
\end{equation}
Additionally, batches for training are created containing as many distinct $V^*$
values as possible. This can be done as the $V^*$ values for the states %$S$
in the training set are pre-computed.

%% file: sections/baselines.tex
\section{Baselines}

We compare the new ``architecture'' \extended{$t$}
with four baseline architectures: (plain) \baseline, Edge Transformers (ETs),
\allcompositions{}, and 2-GNNs. While ETs and \allcompositions{} aim to match the expressive
power of 2-FWL and hence $C_3$, 2-GNNs match the expressive power of 2-OWL and hence $C_2$,
but embedding also pairs of objects. 3-GNNs are unfeasible in practice given the large number
of objects; e.g., with 50 objects, there are 125,000 object triplets.

\subsection{Edge Transformers (ETs)}

Briefly, ETs are designed to operate on a \emph{complete graph} with $n$ nodes and $n^2$ directed edges~\cite{bergen-et-al-neurips2021}.
Each edge is embedded as a $k$-dimensional feature vector. The core computation of an ET layer is described by the equation:
\begin{equation}
  \bm{f}_{i+1}(u,v) = \mlp{}(\norm{}(\bm{f}_{i}(u,v) + \triatt{}(\norm{}(\bm{f}_{i}(u,v)))))
\end{equation}
where $\norm{}$ represents layer normalization, and $\bm{f}_{i}(u,v)$ is the feature vector for pair $\tup{u,v}$ at layer $i$.
A fundamental aspect of ETs is the triangular attention mechanism.
This mechanism functions by aggregating information from all pairs of edges that share a common vertex;
i.e., for \tup{u,v}, it aggregates information from the pairs $\set{  (\tup{u,w},\tup{w,v}) \,{\mid}\, w\,{\in}\,V}$, like 2-FWL, using a self-attention-based combination of contributions.
The expressive power of ETs is the same as for 2-FWL \cite{muller-et-al-neurips2024-transformers}.

Note that, \edgetransformer{}s requires that information about true atoms in a state $S$ and goal $G$ to be included in the initial embeddings.
This restricts \edgetransformer{} to binary relations, and unary predicates are mapped to binary ones by repeating the first term.
To encode the state and the goal, we learn two $k$-dimensional vectors, $\bm{e}_{p}$ and $\bm{e}_{p_g}$, for each predicate $p$.
The initial embedding $\bm{e}_{o,o'}$ for the pair $\langle o, o' \rangle$ is:
\begin{equation}
  \bm{e}_{o, o'} = \textstyle\sum_{p} \bigl( \bm{e}_{p}\,{\cdot}\,\bracket{\,p(o, o') \in S\,} + \bm{e}_{p_g}\,{\cdot}\,\bracket{\,p(o, o') \in G\,} \bigr),
\end{equation}
where $\bracket{\cdot}$ is the Iverson bracket.

These initial embeddings encode the state and the goal in a way that is suitable for the network.
For the value function, we use the same readout function as for \extended{$t$} in all baselines, which aggregates the final embeddings for pairs of identical objects and feeds the resulting vector into an MLP; cf.\ \eqref{eq:final}.
Hence, the number of embeddings aggregated in the final readout is the same for \baseline, \extended{$t$}, and \edgetransformer{}.

\subsection{\allcompositions{}}

The final two baselines are our own.
Rather than selectively adding $\triangle$ atoms, as in $\extended{t}$, we introduce these atoms based on the object pairs rather than the state and the goal.
For this baseline, we aim to emulate 2-FWL by augmenting the input with all possible atoms of the form:
\begin{equation}
  \triangle(\langle o, o' \rangle, \langle o', o'' \rangle, \langle o, o'' \rangle),
\end{equation}
where each atom encodes a composition of two object pairs.
This approach is similar to \extended{$t$}, but also \edgetransformer{}.
In all three, expressivity beyond $C_2$ is achieved through a ``triangulation'' mechanism specifically designed to mirror the 2-FWL procedure.
In WL terms, both ETs and \allcompositions{} maintain the ``context'' of triplets, thereby achieving greater expressiveness.

\subsection{2-GNNs}

The other baseline is designed to emulate 2-OWL instead of 2-FWL.
This allows us to test whether the performance improvement is due to
the object pairs themselves rather than to the additional atoms.
In 2-OWL, the color $C_{i+1}(\tuple{u,v})$ of a pair $\tuple{u,v}$
is determined based on the color $C_i(\tuple{u,v})$ and the multisets
$\multiset{C_i(\tuple{w,v}) \,{\mid}\, w\,{\in}\,V}$ and $\multiset{C_i(\tuple{u,w}) \,{\mid}\, w\,{\in}\,V}$.
This is emulated by introducing two binary predicates, $p_1$ and $p_2$
to represent these multisets.
The ground atoms that determine which pair communicates are all possible
atoms with one of the following forms:
\begin{align}
  p_1(\tuple{w,v},\tuple{u,v}) \\
  p_2(\tuple{u,w},\tuple{u,v})
\end{align}
The initial embeddings encode the state and goal using the same approach
described for the \edgetransformer{} baseline.

%% file: sections/example.tex
\section{Example: Grid Navigation with Obstacles}

\begin{figure}[t]
  \centering
  \begin{tabular}{cc}
    \resizebox{!}{0.222\linewidth}{
    \begin{tikzpicture}[]
        \draw[very thick] (0.00, 0.00) rectangle (8, 4);
        \draw[thick] (0.00, 0.00) grid (8, 4);
        \draw[thick,fill=green] (0, 0) rectangle (1, 1);
        \draw[thick,fill] (1, 0) rectangle (2, 1);
        \draw[thick,fill] (4, 0) rectangle (5, 1);
        \draw[thick,fill=white] (0, 1) rectangle (1, 2);
        \draw[thick,fill=white] (1, 1) rectangle (2, 2);
        \draw[thick,fill] (2, 1) rectangle (3, 2);
        \draw[thick,fill] (3, 1) rectangle (4, 2);
        \draw[thick,fill] (5, 1) rectangle (6, 2);
        \draw[thick,fill] (0, 2) rectangle (1, 3);
        \node (robot) at (1.5, 2.5) {\includegraphics[width=.10\linewidth]{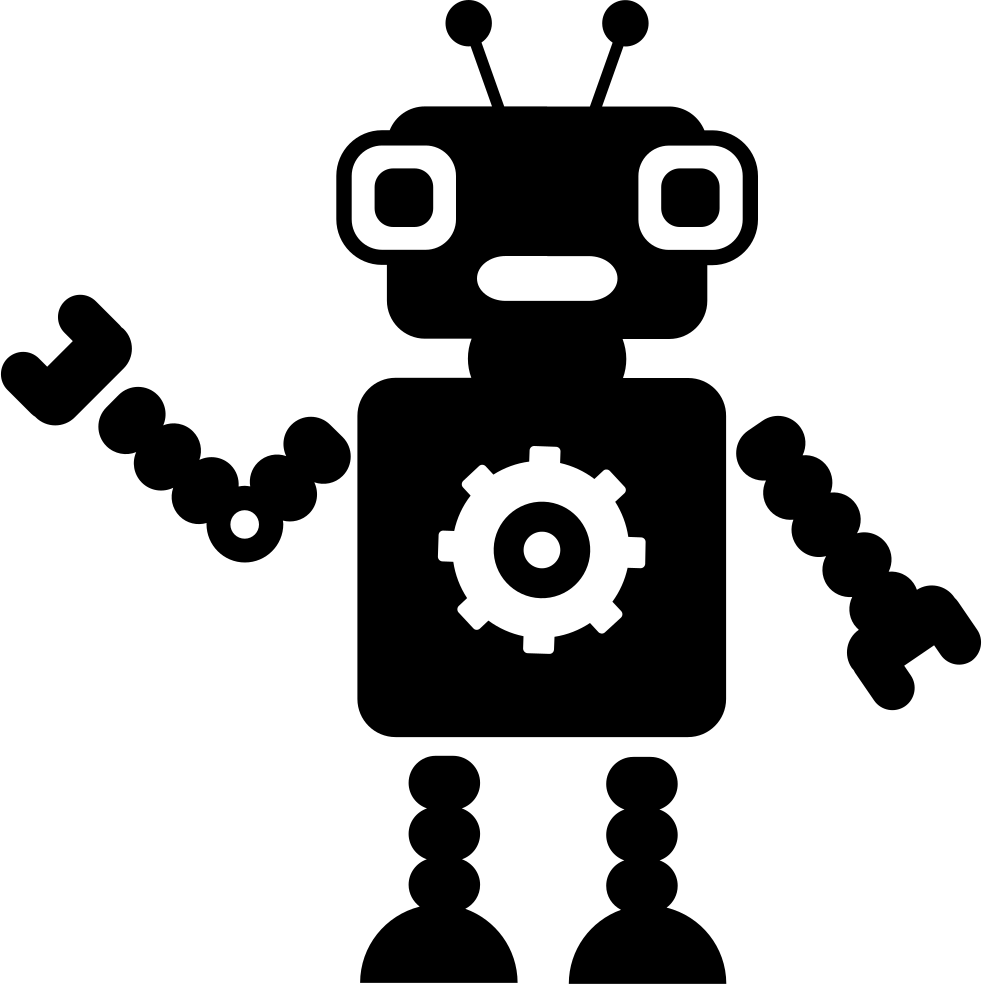}};
        \draw[thick,fill] (2, 2) rectangle (3, 3);
        \draw[thick,fill] (3, 2) rectangle (4, 3);
        \draw[thick,fill] (5, 2) rectangle (6, 3);
        \draw[thick,fill] (6, 2) rectangle (7, 3);
        \draw[thick,fill] (7, 2) rectangle (8, 3);
        \draw[thick,fill] (0, 3) rectangle (1, 4);
        \draw[thick,fill] (1, 3) rectangle (2, 4);
        \draw[thick,fill] (4, 3) rectangle (5, 4);
        \draw[thick,fill] (5, 3) rectangle (6, 4);
        \draw[thick,fill] (7, 3) rectangle (8, 4);
    \end{tikzpicture}
    }
    &
    \resizebox{!}{0.222\linewidth}{
    \begin{tikzpicture}[]
        \draw[very thick] (0.00, 0.00) rectangle (8, 4);
        \draw[thick] (0.00, 0.00) grid (8, 4);
        \draw[thick,fill] (1, 0) rectangle (2, 1);
        \draw[thick,fill] (2, 0) rectangle (3, 1);
        \draw[thick,fill] (3, 0) rectangle (4, 1);
        \draw[thick,fill=white] (4, 0) rectangle (5, 1);
        \draw[thick,fill=white] (5, 0) rectangle (6, 1);
        \draw[thick,fill=white] (6, 0) rectangle (7, 1);
        \draw[thick,fill] (7, 0) rectangle (8, 1);
        \draw[thick,fill] (0, 1) rectangle (1, 2);
        \draw[thick,fill] (1, 1) rectangle (2, 2);
        \draw[thick,fill] (2, 1) rectangle (3, 2);
        \draw[thick,fill=green] (3, 1) rectangle (4, 2);
        \draw[thick,fill=white] (4, 1) rectangle (5, 2);
        \draw[thick,fill] (5, 1) rectangle (6, 2);
        \draw[thick,fill=white] (6, 1) rectangle (7, 2);
        \draw[thick,fill] (7, 1) rectangle (8, 2);
        \node (robot) at (0.5, 2.5) {\includegraphics[width=.10\linewidth]{images/robot2}};
        \draw[thick,fill=white] (1, 2) rectangle (2, 3);
        \draw[thick,fill=white] (2, 2) rectangle (3, 3);
        \draw[thick,fill] (3, 2) rectangle (4, 3);
        \draw[thick,fill] (4, 2) rectangle (5, 3);
        \draw[thick,fill] (5, 2) rectangle (6, 3);
        \draw[thick,fill=white] (6, 2) rectangle (7, 3);
        \draw[thick,fill] (7, 2) rectangle (8, 3);
        \draw[thick,fill] (0, 3) rectangle (1, 4);
        \draw[thick,fill] (1, 3) rectangle (2, 4);
        \draw[thick,fill=white] (2, 3) rectangle (3, 4);
        \draw[thick,fill=white] (3, 3) rectangle (4, 4);
        \draw[thick,fill=white] (4, 3) rectangle (5, 4);
        \draw[thick,fill=white] (5, 3) rectangle (6, 4);
        \draw[thick,fill=white] (6, 3) rectangle (7, 4);
        \draw[thick,fill] (7, 3) rectangle (8, 4);
    \end{tikzpicture}
    }
  \end{tabular}
  \caption{Two $8{\times}4$ test instances of the Navig-xy domain where the
    robot has to  reach the green cell in a grid with obstacles, and  where the
    objects are  the values of each one of the two coordinates, and not the
    cells themselves. The problem is not in $C_2$ in this representation,  and
    indeed, after training, the  baseline \baseline solves the instance on the
    left but not the one on the right, while \extended{$t$}, for $t\,{=}\,1$,
    solves all the instances in the test set. }
  \label{fig:obstacles}
\end{figure}

We illustrate the expressivity demands on a simple example and how these demands
are met by the different architectures. We call the domain Navig-xy and two
instances are shown in Figure~\ref{fig:obstacles}. In this domain, a robot has
to reach the goal (green) cell in a $n \times m$ grid with blocked cells, that
is represented with $n+m$ objects in $X\,{\cup}\,Y$, where
$X\,{=}\,\set{x_1,\ldots,x_n}$ and $Y\,{=}\,\set{y_1,\ldots,y_m}$, and successor
relations $\textsc{Succ-x}$ and $\textsc{Succ-y}$. The domain also includes  the
binary relations $\textsc{At}(x,y)$ to specify the initial and goal cells,
$\textsc{Blocked}(x,y)$ to specify the  cells that are blocked, and a dummy
$\textsc{Cell}(x,y)$ to identify the cells in the grid, for $x \in X$ and $y \in
Y$.

Under the experimental settings described below and 12 hours of training over
105 random $n \times m$ solvable instances, with $nm < 30$, policies strictly
greedy in the learned value function $V$ achieve coverages of 59.72\%, 80.55\%,
and 100\% for the baseline \baseline, \extended{0}, and \extended{1},
respectively, on instances with different sets of blocked cells and up to a
slightly larger size $nm\leq 32$. The ET performs poorly and achieves 4.16\% of
coverage. The instances are not difficult as there is just one free path to
the goal on which the robot just has to keep moving forward, but when the value
function is wrong, it can drive the robot backwards creating a cycle.

The explanation for the coverage results is simple. For computing the true
distance to the goal in these grids, emulating a shortest path algorithm, each
cell $(x,y)$ in the grid must be able to communicate with each of its neighbor
cells  $(x,y')$ and $(x',y)$. In the R-GNN architecture captured by
Alg.~\ref{alg:relnn}, this means that there must be atoms involving the three
objects $x$, $y$, and $y'$, and similarly, $x$, $x'$, and $y$. There are no such
atoms in the state $S$, except in \extended{$t$}, for $t\geq 1$, where $A_t(S)$
includes the composition atoms $\join((x,x'),(x',y),(x,y))$  and
$\join((x,y),(y,y'),(x,y'))$. As a result, \extended{1} and \extended{2} can
compute the true distances, while \extended{0} and \baseline can only compute
``Manhattan distances'', that in some cases (e.g., the left grid in
Fig.~\ref{fig:obstacles}) are good or perfect approximations of $V^*(s)$.

%% file: sections/experiments.tex
\section{Experiments}

A learned value function $V$, for a domain, defines a general policy $\pi_V$ that at state $S$ selects an \emph{unvisited} successor state $S'$ with lowest $V(S')$ value.
We test such policies on instances until reaching a goal state, executing $1000$ steps, or reaching a state with no unvisited successors.
Reaching a goal is counted as a success, else as a failure.

For learning value functions, we implemented the architectures in PyTorch,
and trained the models on NVIDIA A10 GPUs with 24 GB of memory over 12 hours,
using Adam \cite{kingma-et-al-iclr2015} with a learning rate of $0.0002$, batches
of size $16$, and without applying any regularization loss.\footnote{Code, data and models: \url{https://zenodo.org/records/14505092}.}
For each domain, a total of three models were trained, and the model with the lowest loss on the validation set was selected as the final model.
We used embedding dimension $k\,{=}\,64$, $L\,{=}\,30$ layers for \baseline{}, \extended{$t$} and the ETs.
For \allcompositions{}, we used $k\,{=}\,32$ to avoid running out of memory during training.
In all approaches, all layers share weights, and the ETs have 8 self-attention heads.

Inference time depends on the size of the net, but it is typically in the order of tens of milliseconds.
The time to decide which successor to take is the number of successor states multiplied by this time.

\input{sections/domains.tex}

\begin{table}[t]
  \centering
  \resizebox{1.\linewidth}{!}{
  \input{sections/tables/coverage_c2.tex}
  }
  \caption{Coverage and plan lengths for $C_2$ domains.
  In these domains, \baseline{}s performs best, and \extended{$1$} is competitive, except in Gripper.}
  \label{tbl:results-c2}
\end{table}
\begin{table}[t]
  \centering
  \resizebox{1.\linewidth}{!}{
  \input{sections/tables/coverage_c3.tex}
  }
  \caption{Coverage and plan lengths for $C_{3}$ domains.
  In these domains, \extended{$1$} performs best, but both \extended{$0$} and \allcompositions{} outperform \baseline{} and ET.}
  \label{tbl:results-c3}
\end{table}

\subsection{Results}

Tables~\ref{tbl:results-c2} and \ref{tbl:results-c3} show the results.
We anticipate that the improved expressiveness of the networks will result in:
\begin{enumerate}[$\bullet$]
  \item Maintaining performance levels on $C_2$ domains; and
  \item Achieving broader coverage on $C_{3+}$ domains, or
  \item Generating plans of superior quality.
\end{enumerate}
These expections are mostly confirmed by the experiments.
The coverage results for \extended{$t=0,1$} on $C_2$ domains, as seen in Table~\ref{tbl:results-c2},
remain consistent with \baseline{}, with plan quality largely unchanged.
Just in one case, Gripper, increasing the parameter $t$ from 0 to 1 leads
to a decline in coverage for \extended{$t$}.
We attribute this to the high volume of messages that are exchanged,
which slows down training and may result in incomplete convergence.
The plan quality across all approaches is comparable for the $C_2$
domains, except in the Gripper domain, where \baseline{} produces
longer plans.
This is not due to a lack of expressiveness; rather, we believe that
the additional expressiveness provided by the other approaches result
in a more stable general policy.

For the $C_3$ domains, shown in Table~\ref{tbl:results-c3}, we observe that \baseline{} has limited coverage across all domains except Vacuum, where it generates very long plans.
The Vacuum domain requires $C_3$ expressiveness, as each robot has its own traversal capabilities, necessitating the network to determine which agent is closest relative to their capabilities.
While \baseline{} achieves high coverage, actions are executed without clear intention, and goal states are reached incidentally.
This is reflected in the long plan lengths for \baseline{}, whereas other approaches produce optimal or near-optimal plans.

In the other $C_3$ domains, \extended{$1$} consistently outperforms \baseline{} in coverage due to its increased expressiveness.
The performance difference between \extended{$0$} and \extended{$1$} depends on the need for composition.
In Grid, Logistics, and Visitall-xy, at least one level of composition is required, and by including these atoms, we observe improved coverage.
In Rovers, although the necessity for composition is unclear, the plan quality is significantly improved.
Optimal planning in Grid~\cite{helmert-aij2003} is NP-hard, and it seems to be challenging in Rovers as well, and this appears to be the reason why less than 100 \% coverage was achieved.

The baseline \allcompositions{} surpasses ET in all domains except Blocks-s.
We believe this is due to the aggregation function: the output of the softmax in the attention mechanism depend on the number of objects, leading to value magnitudes that differ from those encountered during training.
This is not an issue in \allcompositions{}, where a smooth maximum is used as an aggregation function.
While \allcompositions{} performs better than \baseline{} in $C_3$ domains, it underperforms compared to \extended{$1$}.
This discrepancy is not due to expressiveness, as \allcompositions{} is theoretically more expressive.
Rather, it may be easier to identify the relevant compositions since \extended{$1$} has far fewer compositions in its input.

The baseline \twognn{} consistently performs worse than \extended{$0$} in our experiments, even though both models use object pairs and do not derive compositions.
This disparity is likely due to the reduced volume of messages passed in \extended{$0$}, which allows for clearer messages.
Additionally, each message in \extended{$0$} is computed using MLPs tailored to the predicate symbols of the atoms, leading to more inductive bias and thus better generalization.

%% file: sections/domains.tex
\subsection*{Domains}

Brief descriptions of the domains used in the experiments, mostly taken from \citeauthor{stahlberg-et-al-icaps2022}~(\citeyear{stahlberg-et-al-icaps2022,stahlberg-et-al-kr2022,stahlberg-et-al-kr2023}), follow.
In all cases, the instances in the training set are small, while those in the test set are significantly larger as they contain more objects.

\paragraph{Blocks.}
In Blocks-s (resp.\ Blocks-m), a single tower (resp.\ multiple towers) must be built.
Both have training and validation sets with 4 to 9 blocks.
The test set for Blocks-s (resp.\ Blocks-m) has 10 to 17 blocks (resp.\ up to 20 blocks).

\paragraph{Grid.}
The goal is to fetch keys and unlock doors to reach a cell.
A generator creates random instances with given layouts.
Test instances usually have more keys and locks than those for training and validation,
have different layouts, and their state spaces are too big to be fully expanded.

\paragraph{Gripper.}
A robot with two grippers must move balls from one to another room.
The training and validation instances have up to 14 balls, while test instances have 16-50 balls.

\paragraph{Logistics.}
Transportation domain with packages, cities, trucks, and one airplane.
Training and validation instances have 2-5 cities and 3-5 packages,
while testing instances have 15-19 cities and 8-11 packages.

\paragraph{Miconic.}
An elevator must pick and deliver passengers at different floors.
Training and validation instances involve 2-20 floors and 1-10 passengers,
while those for testing contain 11-30 floors and 22-60 passengers.

\paragraph{Rovers.}
The domain simulates planetary missions where a rover must travel to collect
soil/rock samples, take pictures, and send information back to base. Training
and validation instances use 2-3 rovers and 3-8 waypoints; those for testing
have 3 rovers and 21-39 waypoints.

\paragraph{Vacuum.}
Robot vacuum cleaners that move around and clean different locations.
The robots have their own traversal map, so some robots can go between
two locations while others cannot.
In our version, there is a single dirty location in the middle.
The training and validation sets involve 8-38 locations and 1-6 robots.
The test set includes 40-93 locations and 6-10 robots.

\paragraph{Visitall.}
A robot must visit multiple cells in a grid without obstacles. In Visitall-xy,
the grid is described with coordinates as in the Navig-xy domain, while in
Visitall there is an object for each cell in the grid. Both versions come with
training and validation sets with up to 21 locations, while the test set
includes strictly more, up to 100 cells.

%% file: sections/tables/coverage_c2.tex
\begin{tabular}{@{\extracolsep{-3pt}}lccccc@{}}
  \toprule
              &                     &                                                       & \multicolumn{3}{c}{Plan Length}                 \\
  \cmidrule{4-6}
  Domain                            & Model               & Coverage (\%)                   & Total                           & Median & Mean \\
  \midrule
  Blocks-s                          & \baseline           & \bf 17 / 17 (100 \%)            & 674                             & 38     & 39   \\
                                    & \extended{$0$}      & \bf 17 / 17 (100 \%)            & 670                             & 36     & 39   \\
                                    & \extended{$1$}      & \bf 17 / 17 (100 \%)            & 684                             & 36     & 40   \\
                                   %& \extended{$2$}      & 13 / 17 (76 \%)                 & 566                             & 34     & 43   \\
                                    & \allcompositions{}  & 14 / 17 (82 \%)                 & 922                             & 35     & 65   \\
                                    & \twognn{}           & \bf 17 / 17 (100 \%)            & 678                             & 36     & 39   \\
                                    & \edgetransformer    & 16 / 17 (94 \%)                 & 826                             & 38     & 51   \\
  \addlinespace
  Blocks-m                          & \baseline           & \bf 22 / 22 (100 \%)            & 868                             & 40     & 39   \\
                                    & \extended{$0$}      & \bf 22 / 22 (100 \%)            & 830                             & 39     & 37   \\
                                    & \extended{$1$}      & \bf 22 / 22 (100 \%)            & 834                             & 39     & 37   \\
                                   %& \extended{$2$}      & \bf 22 / 22 (100 \%)            & 824                             & 39     & 37   \\
                                    & \allcompositions{}  & \bf 22 / 22 (100 \%)            & 936                             & 39     & 42   \\
                                    & \twognn{}           & 20 / 22 (91 \%)                 & 750                             & 40     & 37   \\
                                    & \edgetransformer    & 18 / 22 (82 \%)                 & 966                             & 39     & 53   \\
  \addlinespace
  Gripper                           & \baseline           & \bf 18 / 18 (100 \%)            & 4,800                           & 231    & 266  \\
                                    & \extended{$0$}      & \bf 18 / 18 (100 \%)            & 1,764                           & 98     & 98   \\
            \textcolor{white}{\bf )}& \extended{$1$}      & 11 / 18 (61 \%)                 & 847                             & 77     & 77   \\
                                   %& \extended{$2$}      & 4 / 18 (22 \%)                  & 878                             & 86     & 219  \\
                                    & \allcompositions{}  & \bf 18 / 18 (100 \%)            & 1,764                           & 98     & 98   \\
                                    & \twognn{}           & 1 / 18 (6 \%)                   & 53                              & 53     & 53   \\
                                    & \edgetransformer    & 4 / 18 (22 \%)                  & 246                             & 61     & 61   \\
  \addlinespace
  Miconic                           & \baseline           & \bf 20 / 20 (100 \%)            & 1,342                           & 67     & 67   \\
                                    & \extended{$0$}      & \bf 20 / 20 (100 \%)            & 1,566                           & 71     & 78   \\
                                    & \extended{$1$}      & \bf 20 / 20 (100 \%)            & 2,576                           & 71     & 128  \\
                                   %& \extended{$2$}      & 19 / 20 (95 \%)                 & 1,697                           & 70     & 89   \\
                                    & \allcompositions{}  & \bf 20 / 20 (100 \%)            & 1,342                           & 67     & 67   \\
                                    & \twognn{}           & 12 / 20 (60 \%)                 & 649                             & 54.5   & 54   \\
                                    & \edgetransformer    & \bf 20 / 20 (100 \%)            & 1,368                           & 68     & 68   \\
  \addlinespace
  Visitall                          & \baseline           & 18 / 22 (82 \%)                 & 636                             & 29     & 35   \\
                                    & \extended{$0$}      & 21 / 22 (95 \%)                 & 1,128                           & 35     & 53   \\
                                    & \extended{$1$}      & \bf 22 / 22 (100 \%)            & 886                             & 35     & 40   \\
                                   %& \extended{$2$}      & \bf 22 / 22 (100 \%)            & 867                             & 34     & 39   \\
                                    & \allcompositions{}  & 20 / 22 (91 \%)                 & 739                             & 33     & 36   \\
                                    & \twognn{}           & 18 / 22 (82 \%)                 & 626                             & 32     & 34   \\
                                    & \edgetransformer    & 18 / 22 (82 \%)                 & 670                             & 29     & 37   \\
  \bottomrule
\end{tabular}

%% file: sections/tables/coverage_c3.tex
\begin{tabular}{@{\extracolsep{-3pt}}lccccc@{}}
  \toprule
              &                    &                                                       & \multicolumn{3}{c}{Plan Length}                 \\
  \cmidrule{4-6}
  Domain      & Model              & Coverage (\%)                                                              & Total                           & Median & Mean \\
  \midrule
  Grid     \textcolor{white}{\bf )}& \baseline          & 9 / 20 (45 \%)                                        & 109                             & 11     & 12   \\
           \textcolor{white}{\bf )}& \extended{$0$}     & 12 / 20 (60 \%)                                       & 177                             & 11     & 14   \\
                                   & \extended{$1$}     & \bf 15 / 20 (75 \%)                                   & 209                             & 13     & 13   \\
                                  %& \extended{$2$}     & \multicolumn{4}{c}{Ran out of memory during training}                                                   \\
                                   & \allcompositions{} & 10 / 20 (50 \%)                                       & 124                             & 11.5   & 12   \\
           \textcolor{white}{\bf )}& \twognn{}          & 6 / 20 (30 \%)                                        & 82                              & 11.5   & 13   \\
                                   & \edgetransformer   & 1 / 20 (5 \%)                                         & 15                              & 15     & 15   \\
  \addlinespace
  Logistics\textcolor{white}{\bf )}& \baseline          & 10 / 20 (50 \%)                                       & 510                             & 51     & 51   \\
           \textcolor{white}{\bf )}& \extended{$0$}     & 9 / 20 (45 \%)                                        & 439                             & 48     & 48   \\
                                   & \extended{$1$}     & \bf 20 / 20 (100 \%)                                  & 1,057                           & 52     & 52   \\
                                  %& \extended{$2$}     & \bf 20 / 20 (100 \%)                                  & 1,057                           & 52     & 52   \\
           \textcolor{white}{\bf )}& \allcompositions{} & 15 / 20 (75 \%)                                       & 799                             & 52     & 53   \\
                                   & \twognn{}          & 0 / 20 (0 \%)                                         & --                              & --     & --   \\
                                   & \edgetransformer   & 0 / 20 (0 \%)                                         & --                              & --     & --   \\
  \addlinespace
  Rovers   \textcolor{white}{\bf )}& \baseline          & 9 / 20 (45 \%)                                        & 2,599                           & 280    & 288  \\
                                   & \extended{$0$}     & \bf 14 / 20 (70 \%)                                   & 2,418                           & 153    & 172  \\
                                   & \extended{$1$}     & \bf 14 / 20 (70 \%)                                   & 1,654                           & 55     & 118  \\
                                  %& \extended{$2$}     & 13 / 20 (65 \%)                                       & 2,600                           & 136    & 200  \\
           \textcolor{white}{\bf )}& \allcompositions{} & 11 / 20 (55 \%)                                       & 2,225                           & 239    & 202  \\
                                   & \twognn{}          & \multicolumn{4}{c}{\multirow{2}{*}{Unsuitable domain: ternary predicates}}                              \\
                                   & \edgetransformer   & \multicolumn{4}{c}{}                                                                                    \\
  \addlinespace
  Vacuum                           & \baseline          & \bf 20 / 20 (100 \%)                                  & 4,317                           & 141    & 215  \\
                                   & \extended{$0$}     & \bf 20 / 20 (100 \%)                                  & 183                             & 9      & 9    \\
                                   & \extended{$1$}     & \bf 20 / 20 (100 \%)                                  & 192                             & 9      & 9    \\
                                  %& \extended{$2$}     & 19 / 20 (95 \%)                                       & 246                             & 10     & 12   \\
                                   & \allcompositions{} & \bf 20 / 20 (100 \%)                                  & 226                             & 9      & 11   \\
                                   & \twognn{}          & \multicolumn{4}{c}{\multirow{2}{*}{Unsuitable domain: ternary predicates}}                              \\
                                   & \edgetransformer   & \multicolumn{4}{c}{}                                                                                    \\[4pt]
  \addlinespace
  Visitall-xy                      & \baseline          & 5 / 20 (25 \%)                                        & 893                             & 166    & 178  \\
                                   & \extended{$0$}     & 15 / 20 (75 \%)                                       & 1,461                           & 84     & 97   \\
                                   & \extended{$1$}     & \bf 20 / 20 (100 \%)                                  & 1,829                           & 83     & 91   \\
                                  %& \extended{$2$}     & 18 / 20 (90 \%)                                       & 1,631                           & 86     & 90   \\
                                   & \allcompositions{} & 19 / 20 (95 \%)                                       & 2,428                           & 116    & 127  \\
                                   & \twognn{}          & 12 / 20 (60 \%)                                       & 1,435                           & 115    & 119  \\
                                   & \edgetransformer   & 3 / 20 (15 \%)                                        & 455                             & 138    & 151  \\
  \bottomrule
\end{tabular}

%% file: sections/expressivity.tex
\section{On the Expressivity of the Model}

We establish next that the \extended{$t$} model has the capability
to capture compositions of binary relations that can be expressed
in $C_3$.
This capability is critical in many domains, and can be achieved
by adding derived predicates \cite{stahlberg-et-al-kr2022,haslum-et-al-2019,thiebaux-et-al-aij2005}.
In particular, we are interested in derived predicates that correspond
to relational joins in $C_3$:

\begin{definition}[$C_3$-Relational Joins]
  Let $\sigma$ be a relational language.
  The class $\J_3=\J_3[\sigma]$ of relational joins over the language $\sigma$
  is the \textbf{smallest} class of formulas that satisfy the following properties:
  \begin{enumerate}[1.]%\topsep0pt%\itemsep0pt
    \item $\set{R(x,y), \neg R(x,y)} \subseteq \J_3$ for binary relation $R$ in $\sigma$,
    \item $\varphi(x,y) \land \phi(x,y) \in \J_3$ if $\set{\varphi(x,y), \phi(x,y)} \subseteq \J_3$,
    \item $\varphi(x,y) \lor \phi(x,y) \in \J_3$ if $\set{\varphi(x,y), \phi(x,y)} \subseteq \J_3$, and
    \item $\exists z[ \varphi(x,z) \land \phi(y,z) ] \in \J_3$ if $\set{\varphi(x,y), \phi(y,z)} \subseteq \J_3$.
  \end{enumerate}
  The notation $\varphi(x,y)$ means that $\varphi$ is a formula whose
  free variables are among $\set{x,y}$.
\end{definition}

For example, if $\sigma$ contains the relations $\textsc{Key}(k,s)$ and $\textsc{Lock}(\ell,t)$
to express that the key $k$ has shape $s$, and the lock $\ell$ has shape $t$, respectively,
then $\J_3$ contains $\varphi(k,\ell) = \exists s[ \textsc{Key}(k,s) \land \textsc{Lock}(\ell,s) ]$
that is true for the pair $\tuple{k,\ell}$ when the key $k$ opens the lock $\ell$.

Let $\STRUC=\STRUC[\sigma]$ be the class of finite structures for language $\sigma$.
A network that maps structures $\A$ in \STRUC into embeddings for all the $k$-tuples
$\tuple{u_1,u_2,\ldots,u_k}$ of objects in $\A$ is called a \textbf{$k$-embedding network,}
and a collection of such networks is a \textbf{$k$-embedding architecture.}
For example, the class of all nets in \extended{$t$} for $\sigma$ is a 2-embedding
architecture.
The architecture $\extended{\sigma,t,k,L}$ is the collection of networks in \extended{$t$}
for the language $\sigma$, embedding dimension $k$, and $L$ layers.

Let $\varphi(x,y)$ be a relational join, and let $\A$ be a structure with universe $U$.
The denotation of $\varphi(x,y)$ over $\A$, denoted by $\A^\varphi$, is the set
of pairs $\set{\tuple{u,v}\in U^2 \mid \A\vDash \varphi(u,v)}$.
A 2-embedding network $N$ with embedding dimension $k$ computes $\varphi(x,y)$
if there is an index $0\leq j<k$ such that the pair $\tuple{u,v}\in\A^\varphi$ iff
$\bm{f}(\tuple{u,v})_j=1$, where $\bm{f}(\tuple{u,v})$ is the embedding for $\tuple{u,v}$
produced by $N$ on input $\A$.
Likewise, such a network $N$ computes a collection $\D$ of relational
joins if $N$ computes each join $\varphi(x,y)$ in $\D$.

\begin{theorem}[Computation of $C_3$-Relational Joins]
  Let $\sigma$ be a relational language, and let $\D$ be
  a \textbf{finite collection} of $C_3$-relational joins.
  Then, there is a tuple of parameters $\tuple{t,k,L}$ and
  network $N$ in $\extended{\sigma,t,k,L}$ that computes $\D$.
\end{theorem}

The parameters $\tuple{t,k,L}$ are determined by the joins in $\D$.
The index $t$ that defines the nesting depth of the $\Delta$ atoms is the
maximum quantifier depth over (the joins in) $\D$.
On the other hand, the embedding dimension $k$ and the number of layers $L$
are bounded by the sum and maximum, respectively, of the number of subformulas
of the joins in $\D$.

To illustrate the capabilities of the proposed architecture, let us consider
the Navig-xy domain from above for which $\sigma$ contains the relations \textsc{Succ-x}, \textsc{Succ-y},
\textsc{At}, $\textsc{At}_g$, \textsc{Blocked}, and \textsc{Cell}.
We want to show that $\J_3$ contains the predicate $\varphi_k(x,y)$ that tells
when the cell \tuple{x,y} is at $k$ steps from the goal cell.
Indeed, $\varphi_0(x,y)=\textsc{At}_g(x,y)$ is in $\J_3$.
Likewise,
\begin{alignat*}{1}
  \textsc{Adj-x}(x,x')\ &=\ \textsc{Succ-x}(x,x') \lor \textsc{Succ-x}(x',x)\,, \\
  \textsc{Adj-y}(y,y')\ &=\ \textsc{Succ-y}(y,y') \lor \textsc{Succ-y}(y',y)
\end{alignat*}
belong to $\J_3$. Then, the following also belong to $\J_3$:
\begin{alignat*}{1}
  \phi^\textsc{x}_{k+1}(x,y)\ &=\ \exists z[\varphi_k(z,y) \land \textsc{Adj-x}(x,z)] \,, \\
  \phi^\textsc{y}_{k+1}(x,y)\ &=\ \exists z[\varphi_k(x,z) \land \textsc{Adj-y}(y,z)] \,, \\
  \phi_{k+1}(x,y)\            &=\ \phi^\textsc{x}_{k+1}(x,y) \lor \phi^\textsc{y}_{k+1}(x,y) \,, \\
  \varphi_{k+1}(x,y)\         &=\ \neg\textsc{Blocked}(x,y) \land \phi_{k+1}(x,y) \,.
\end{alignat*}
Hence, for any state $S$ in an instance of Navig-xy, $V^*(S)=k$ iff $S\vDash\textsc{Dist}_k$ where
$\textsc{Dist}_k = \exists xy[ \textsc{At}(x,y) \land \varphi_k(x,y)]$ is a sentence in $\J_3$.
Therefore, for a class of instances of bounded $x$ and $y$ dimensions, there is an embedding
dimension $k$, a number $L$ of layers, and a network $N$ in \extended{$\sigma,1,k,L$} that
computes $V^*$ for the states $S$ in such instances.

%% file: sections/conclusions.tex
\section{Conclusions}

The paper presents a novel approach for extending the expressive power of Relational Graph Neural Networks (R-GNNs) in the classical planning setting by just adding a set of atoms $A_t$ to the state, in a domain-independent manner that depends on the $t$ parameter and the pairs of objects that interact in an atom true in the state.
The resulting ``architecture'' R-GNN[$t$] appears to produce the necessary $C_3$ features in a practical manner, without the memory and time overhead of 3-GNNs, and with much better generalization ability than Edge Transformers, which have the expressiveness of $3$-GNNs, but with much less overhead.
Interestingly, for all of the domains we considered, we did not see any improvement using $t > 1$.
This is exploited by \extended{$t$}, as far fewer compositions need to be considered, resulting in much faster inference and models that generalize better than the baselines that consider all possible compositions.

It remains to be studied whether the full power of $C_3$ is needed to handle most planning domains.
The expressivity results from \citeay{horcik-sir-icaps2024} and \citeay{drexler-et-al-kr2024} suggest that the expressive power of $C_3$ is sufficient but not necessary.
Indeed, there is ample room in the middle, between $C_2$ and $C_3$, which the \extended{$t$} architecture exploits, in contrast with \allcompositions{} and ETs.
Approaches that aim to extend representations with new unary predicates by considering cycles in the state graphs may yield an acceptable tradeoff between expressivity and efficiency, without the need to embed pairs or higher tuples of objects at all.
This is left for future work.

%% file: sections/acknowledgements.tex
\section*{Acknowledgments}

The research has been supported by the Alexander von Humboldt Foundation with funds from the Federal Ministry for Education and Research, Germany, by the European Research Council (ERC), Grant agreement No. 885107, by the Excellence Strategy of the Federal Government and the NRW L\"{a}nder, Germany, and by the Knut and Alice Wallenberg (KAW) Foundation under the WASP program.